\documentclass[sigconf]{acmart}

\setcopyright{none}
\renewcommand\footnotetextcopyrightpermission[1]{} 
\AtBeginDocument{%
  \providecommand\BibTeX{{%
    \normalfont B\kern-0.5em{\scshape i\kern-0.25em b}\kern-0.8em\TeX}}}

\copyrightyear{2023}
\acmYear{2023}
\setcopyright{rightsretained}
\acmConference[Arxiv]{Arxiv}{2023}{}
\acmBooktitle{Arxiv}
\acmDOI{10}
\acmISBN{10}

\usepackage{comment}
\usepackage{algorithm}
\usepackage{algorithmic}

\DeclareMathOperator{\Unif}{Unif}

\author{Zimin Liang}
\affiliation{%
  \institution{University of Birmingham}
  \city{Birmingham}
  \country{UK}
}
\email{z.liang.1@bham.ac.uk}

\author{Miqing Li}
\authornote{Corresponding author}
\affiliation{%
  \institution{University of Birmingham}
  \city{Birmingham}
  \country{UK}
}
\email{m.li.8@bham.ac.uk}

\author{Per Kristian Lehre}
\affiliation{%
  \institution{University of Birmingham}
  \city{Birmingham}
  \country{UK}
}
\email{p.k.lehre@bham.ac.uk}

\begin{document}

\title{Non-Elitist Evolutionary Multi-Objective Optimisation: Proof-of-Principle Results}




\begin{abstract}
 Elitism, which constructs the new population by preserving best solutions out of the old population and newly-generated solutions,
 has been a default way for population update since its introduction into multi-objective evolutionary algorithms (MOEAs) in the late 1990s.
In this paper, we take an opposite perspective to conduct the population update in MOEAs by simply discarding elitism.
That is,
we treat the newly-generated solutions as the new population directly
(so that all selection pressure comes from mating selection).
We propose a simple non-elitist MOEA (called NE-MOEA) that only uses Pareto dominance sorting to compare solutions, without involving any diversity-related selection criterion.
Preliminary experimental results show that NE-MOEA can compete with well-known elitist MOEAs (NSGA-II, SMS-EMOA and NSGA-III) on several combinatorial problems.
Lastly, we discuss limitations of the proposed non-elitist algorithm and suggest possible future research directions. 
 
\end{abstract}



\maketitle

\section{Introduction}

Multiobjective Optimisation problems (MOPs) refer to problems that optimise more than one objective simultaneously. 
A prominent feature of MOPs is that there typically does not exist a single optimal solution but rather a set of trade-off solutions, 
called Pareto optimal solutions or Pareto front in the objective space.
Evolutionary algorithms (EAs) have been demonstrated well-suited to MOPs.
Their population-based search can approximate an MOP's Pareto front in one execution, 
with each individual representing a different trade-off between the objectives. 

In the design of an EA, 
there are two key components, 
solution generation and population update (also known as environmental selection, or population maintenance). 
The former is concerned with how to generate solutions, 
including mating selection and variation.
The latter is concerned with how to update the population based on the old population and newly-generated solutions.

As for the population update, 
the aim is to select well-distributed nondominated solutions for next-generation evolution. 
It is natural that one would like to select the very best solutions from the collections of the old population and newly-generated solutions to form the new population, 
in the hope of a higher chance to generate better solutions next time.
Indeed, since the introduction of SPEA by Zitzler \emph{et al.} in 1999~\cite{zitzler_multiobjective_1999}, 
almost all MOEAs are based on elitism, no matter what selection criteria they used (e.g., the Pareto-based, the decomposition-based or the indicator-based)~\cite{emmerich_tutorial_2018}.

In this paper, 
we take an opposite perspective to conduct the population update in MOEAs, 
by simply dropping elitist preservation. 
That is, 
we discard the old population completely and always treat the newly-generated solutions as the new population. 
The idea is inspired by the success of non-elitist evolutionary algorithms in single-objective optimisation in 
preventing populations from getting stuck in local optima \cite{schaefer_negative_2010, lehre_fitness-levels_2011, dang_runtime_2016}. 
We show the proposed algorithm, 
called non-elitist multiobjective evolutionary algorithm (NE-MOEA),
can outperform the elitist MOEAs on two classes of combinatorial optimisation problems, knapsack problems and NK-Landscape problems,
 under appropriate settings of genetic parameters.

The rest of the paper is organised as follows. 
Section II introduces the preliminaries of multi-objective optimisation and combinatorial problems we consider in this study. 
Section III presents the NE-MOEA framework. 
Section IV is devoted to experimental design.
Section V presents the experimental result and discussion.
Lastly, 
Section VII concludes the paper.

\vspace{-6pt}
\section{Preliminaries}
\subsection{Terminology}

Let us consider a maximisation problem with $n$ decision variables and $m$ objective functions 
$f : X \rightarrow Z$, with $X \subseteq \{0,1\}^n$ and $Z \subseteq \mathbb{R}^m$. 
The objective functions map a binary vector $x \in X$ in the decision space to an objective vector
$z \in Z$ in the objective space such that $z=f(x) = (f_1(x),...,f_m(x))$.

Given two objective vectors $z, z'\in Z$, 
$z$ is said to \emph{dominate} $z'$ (denoted by $z \succ z'$) iff for all $i\in \{1,...,m\}$, $z_i \geq z'_i$, 
and there exists a $j \in \{1,...,m\}$ such that $z_j > z'_j$.
Likewise,
given two solutions $x, x'\in X$,
$x$ is said to dominate $x'$ iff $f(x)$ dominates $f(x')$.
An objective vector $z$ is called \emph{Pareto optimal} if there does not exist any $z'\in Z$ such that $z' \succ z$.
A solution $x \in X$ is Pareto optimal if $f(x)$ is Pareto optimal. 
The set of Pareto optimal solutions is called the \emph{Pareto optimal set}, 
and its mapping in the objective space is called the \emph{Pareto front}.

\subsection{Optimisation Problems}

\vspace{5pt}
\noindent
\textbf{Multi-objective 0/1 knapsack problem~\cite{zitzler_multiobjective_1999}}. 
Multi-objective knapsack problem is to select a subset of items, with each item having multiple values (objectives) and weights (constraints),
in order to maximise their total value on these objectives but at the same time subject to the weight constraints.

\vspace{5pt}
\noindent
\textbf{Multi-objective NK-landscape problem ~\cite{aguirre_insights_2004}}. 
NK-Landscape is a commonly used problem in multi-objective optimisation due to the controllableness of ruggedness of the problem's landscape~\cite{verel_structure_2013, aguirre_insights_2004}. 
In the NK-landscape problem, 
$N$ represents the number of bits (i.e., decision variables),
and $K$ represents the number of bits that affects a bit.

\begin{algorithm}[tbp]
\caption{Elitist MOEA Framework}
\label{alg:moea}
\begin{algorithmic}[1]
\STATE $P_0 \gets \textit{Initialisation()}$
\WHILE{not stopCondition} 
\STATE $Q_t \gets \textit{Reproduction}(P_t)$
\STATE \textcolor{blue}{$P_{t+1}\gets \textit{PopulationUpdate}(P_t,Q_t)$}
\ENDWHILE
\end{algorithmic}
\end{algorithm}
\begin{algorithm}[tbp]
\caption{Non-elitist MOEA Framework}
\label{alg:nemoea}
\begin{algorithmic}[1]
\STATE $P_0 \gets \textit{Initialisation()}$
\WHILE{not stopCondition} 
\STATE $Q_t \gets \textit{Reproduction}(P_t)$
\STATE \textcolor{blue}{$P_{t+1}\gets Q_t$ }
\ENDWHILE
\end{algorithmic}
\end{algorithm}

\section{The Proposed Non-Elitist MOEA}

\subsection{Framework}

Elitist and non-elitist MOEAs differ in the population update part of the evolutionary algorithm. 
Algorithms~\ref{alg:moea} and \ref{alg:nemoea} give the framework of elitist and non-elitist MOEAs, respectively. 
As shown,
after generating new solutions by the old population $P_t$ (line~3),
elitist MOEAs form the new population $P_{t+1}$ by using both the old population $P_t$ and the newly-generated solutions $Q_t$ (line 4 in Algorithm~\ref{alg:moea}),
whereas non-elitist MOEAs treat newly-generated solutions $Q_t$ as the new population $P_{t+1}$ (line 4 in Algorithm~\ref{alg:nemoea}). 
Next, we describe the reproduction part of the proposed non-elitist MOEA.

\subsection{Reproduction}
Reproduction is a process of generating new solutions, through selecting good solutions of the current population (based on their fitness) to perform variation. 
It typically consists of two parts: mating selection and variation (e.g., crossover and mutation).
Since it is a proof-of-principle study, 
we aim to keep the proposed non-elitist MOEA as simple as possible. 
We do not consider the diversity of solutions (but only Pareto dominance) in their fitness for mating selection.
Moreover, for variation we do not consider crossover (but only mutation), which is in line with non-elitist single-objective EAs~\cite{schaefer_negative_2010, dang_non-elitist_2021}.
Algorithm~\ref{alg:reproduction} gives the reproduction procedure of the proposed algorithm.

\renewcommand{\algorithmiccomment}[1]{$/^{*}$ #1 $^{*}/$ }
\begin{algorithm}[tbp]
\caption{Reproduction($P$)}
\label{alg:reproduction}
\begin{algorithmic}[1]
\REQUIRE{$N\in\mathbb{N}$ (popoulation size), $P = (\textbf{x}_1, \ldots, \textbf{x}_N)\in\mathcal{X}^N$ (current population); $k\in\mathbb{N}$ (tournament size)}; 
\STATE $(f(\textbf{x}_1),\ldots,f(\textbf{x}_N)) \gets \textit{NondominatedSorting}(P)$ \\
\COMMENT{assign solutions' fitness by setting it to be the index of the front that the solution is located in after the nondominated sorting procedure~\cite{goldberg_genetic_1989}} 
\FOR{$i = 1 \dots N$}
\FOR{$j=1 \dots k$}
\STATE $\textbf{y}_j \sim \Unif(\textbf{x}_1, \ldots, \textbf{x}_N)$
\quad \quad \COMMENT{draw a solution out of the population uniformly at random}
\ENDFOR
\STATE $\textbf{y}'_i \gets \arg\min_{\textbf{y} \in (\textbf{y}_1, \ldots, \textbf{y}_k)}f(\textbf{y})$ \quad \COMMENT{break ties randomly}
\STATE $\textbf{x}'_i=\textit{Mutation}(\textbf{y}'_i)$
\ENDFOR
\STATE \textbf{return} $Q=(\textbf{x}'_1, \ldots, \textbf{x}'_N)$
\end{algorithmic}
\end{algorithm}

\vspace{5pt}
\noindent{\textbf{Mating Selection.}} 
Mating selection part selects solutions from the current population as parent to perform variation.
Selection pressure in non-elitist MOEAs only comes from the mating selection part and sufficient selection pressure is needed to drive the search towards the Pareto front. 
To do so,
we use $k$-tournament selection (lines 3--6 in Algorithm~\ref{alg:reproduction}) to select the best solution (according to their fitness) out of eight randomly sampled solutions from the current population (i.e., $k=8$), where tie is broken randomly.
For fitness assignment,
we consider a commonly-used method, non-dominated rank computed by fast non-dominated sorting~\cite{deb_fast_2002}.

\vspace{5pt}
\noindent{\textbf{Mutation.}}\label{Sec:mutation}
The proposed non-elitist MOEA does not restrict itself to a specific mutation operator. 
g to the underlying optimisation problem.
However, 
in non-elitist EAs, 
the performance is sensitive to the mutation rate. 
According to \cite{dang_non-elitist_2021, qin_self-adaptation_2022}, 
the mutation rate should be close to but slightly less than the error threshold (see Eq. (\ref{eq:recommended-mut-rate}) where k is the tournament size in mating selection), otherwise the population may not converge. 
\begin{align}
\frac{(1-\delta)\ln(k)}{n}\label{eq:recommended-mut-rate}
\end{align} 
Since runtime analyses for non-elitist MOEAs are currently not available, we follow the setting of single-objective non-elitist EAs, and use mutation rate of $0.95\frac{\ln{k}}{n}$.


\section{Experimental Design}

We consider three representative elitist MOEAs with different selection criteria to compare with our non-elitist MOEA. They are NSGA-II~\cite{deb_fast_2002} (Pareto-based criterion), SMS-EMOA~\cite{beume_sms-emoa_2007} (indicator-based criterion), and NSGA-III~\cite{deb_evolutionary_2014} (decomposition-based criterion).

We use the indicator hypervolume \cite{zitzler_multiobjective_1999} to compare the quality of solution sets generated by the considered algorithms (i.e., their archive). 
In the calculation of hypervolume,
the reference point was set to $(0,0)$.

Non-elitist EAs normally need a large population and many generations~\cite{schaefer_negative_2010,lehre_fitness-levels_2011}.
Here, the size of the population was set to 10,000 and the number of generations was 5,000. 
Thus the total number of evaluations is $5\times 10^7$.
The same settings are used for NSGA-II, SMS-EMOA, and NSGA-III. 
For variation operators, 
we use the bit-flip mutation and uniform crossover in this study.
As for the variation rate,
in the proposed NE-MOEA, 
there is no crossover and the mutation rate is $0.95\frac{\ln{k}}{n}$ (Section~\ref{Sec:mutation}).
For the elitist MOEAs,
we followed a practice widely used (e.g., in~\cite{deb_fast_2002}). 
The crossover rate is 0.9 and the mutation rate is $1/n$, 
where $n$ denotes the number of decision variables. 

Note that in the non-elitist MOEA, 
since the old population is directly discarded,
the population cannot represent the best solutions found so far.
We used an external unbounded archive to store all nondominated solutions generated during the evolutionary process. 
For a fair comparison,
the elitist MOEAs used for comparison in our experiments also used an unbounded archive to store all nondominated solutions generated during the evolutionary process. 
Note that even with elitism, 
the population of MOEAs can deteriorate, i.e., the algorithms may preserve globally dominated solutions while discarding true non-dominated solutions~\cite{Li2019a}.

\begin{table*}[tbp]
    \caption{Hypervolume results (mean and stardard deviation) of NSGA-II, SMS-EMOA, NSGA-III and the proposed NE-MOEA on the bi-objective 0/1-Knapsack problem (KP) and NK-landscape problem (NK), where $n$ in KP is the number of items. For each problem instance, the best mean is highlighted in bold.}
    \begin{tabular}{l|c|c|c|c}
        \textbf{Problems} & \textbf{NSGA-II}  & \textbf{SMS-EMOA}&\textbf{NSGA-III} & \textbf{NE-MOEA} \\ \hline
        KP (n=50)  & \textbf{9.9528e-1} (5.30e-6) & 9.9524e-1 (8.62e-4) 
 & \textbf{9.9528e-1 }(5.99e-6) & \textbf{9.9528e-1 }(0.00e+0) \\ 
        KP (n=100) & 9.8203e-1 (3.01e-3)$\dagger$ & 9.7651e-1 (4.22e-3)$\dagger$ 
 & 9.7204e-1 (5.21e-3)$\dagger$ & \textbf{9.9248e-1 }(7.12e-4) \\
        KP (n=200) & 9.7151e-1 (2.06e-3)$\dagger$ & 9.7057e-1 (4.02e-3)$\dagger$
 & 9.6787e-1 (4.42e-3)$\dagger$ & \textbf{9.8602e-1 }(1.17e-3) \\
        KP (n=300) & 9.7072e-1 (2.05e-3)$\dagger$ & 9.5560e-1 (4.89e-3)$\dagger$
 & 9.5957e-1 (3.63e-3)$\dagger$ & \textbf{9.7919e-1 }(1.19e-3) \\ \hline
        NK (n=50, k=10) & 5.4398e-1 (8.40e-3)$\dagger$ & 5.0616e-1 (1.29e-2)$\dagger$ & 5.5107e-1 (7.74e-3)$\dagger$ & \textbf{5.7815e-1 }(8.30e-3) \\
        NK (n=100, k=10) & 4.5734e-1 (1.70e-2)$\dagger$ & 4.4387e-1 (2.44e-2)$\dagger$ & 4.6097e-1 (1.89e-2)$\dagger$ & \textbf{5.6711e-1 }(6.63e-3) \\
        NK (n=200, k=10) & 4.2847e-1 (1.96e-2)$\dagger$ & 
        4.2571e-1 (2.12e-2)$\dagger$
        & 4.2469e-1 (2.41e-2)$\dagger$ & \textbf{5.5463e-1 }(6.16e-3) \\
        NK (n=300, k=10) & 4.2100e-1 (2.15e-2)$\dagger$ & 3.8225e-1 (2.17e-2)$\dagger$ & 4.0983e-1 (2.37e-2)$\dagger$ 
        & \textbf{5.4785e-1 }(5.32e-3) \\ \hline
    \end{tabular}
    \label{tab:result}
    
    \small
    ``$\dagger$'' indicates that the result is significantly different from that obtained by NE-MOEA at a 95\% confidence by the Wilcoxon rank-sum test.
\end{table*}

\begin{figure}[tbp]
    \centering
    \includegraphics[scale=0.59]{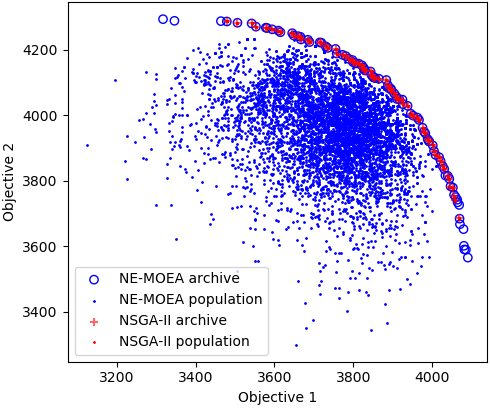}
    \caption{\textbf{Knapsack (100 items)}: 
    \small{The archive and the final population obtained by NE-MOEA and NSGA-II on the 0/1 knapsack problem with 100 items. For NSGA-II, the final population and the archive are virtually overlapping (but not the case for NE-MOEA).}
    }
    \label{fig:kp100}
\end{figure}

\section{Result}

Table \ref{tab:result} gives the hypervolume results obtained by all the four algorithms. 
All results presented throughout the paper are based on 30 independent runs of each comparative algorithm on each problem instance. 
We adopted Wilcoxon's rank-sum (95\% confidence) test to examine the statistical significance of the results.

\vspace{5pt}
\noindent{\textbf{0/1-Knapsack Problem.}}
For the 0/1-Knapsack problem, 
as shown in Table~\ref{tab:result}, 
NSGA-II, NSGA-III and NE-MOEA obtain an identical hypervolume value on the problem with 50 items,
indicating that they perform very similarly. 
As the problem size increases (100, 200 and 300 items), 
NE-MOEA outperforms other elitist algorithms significantly, while NSGA-II takes the second place on most of the problem instances.

Figure~\ref{fig:kp100} gives the solutions (both the final population and external archive) found by NE-MOEA and NSGA-II in a typical run on the 0/1-Knapsack problem with 100 items.
The archives (i.e., those represented by circles and pluses) only store nondominated solutions.
As can be seen from the figure, 
most non-dominated solutions obtained by NSGA-II and NE-MOEA are overlapping,
but NE-MOEA can find some corner solutions whereas NSGA-II fails to.
This enables NE-MOEA to have a better hypervolume value.

\begin{figure}[!]
    \centering
    \includegraphics[scale=0.6]{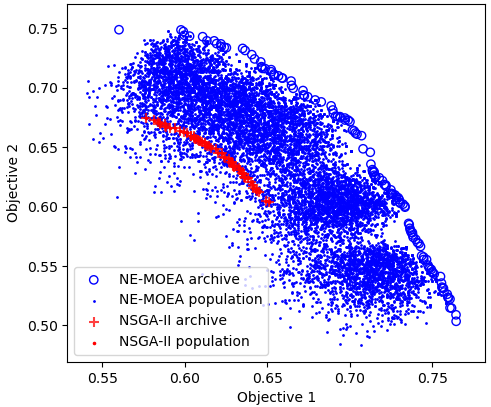}
    \caption{\textbf{NK-Landscape (n=200, k=10)}: 
    \small{The archive and the final population obtained by NE-MOEA and NSGA-II on the NK-Landscape with $n=200$ and $k=10$.}
    }
    \label{fig:nk20010}
\end{figure}

\vspace{5pt}
\noindent{\textbf{NK-Landscape Problem.}}
The NK-Landscape problem is known for having a rugged multi-modal fitness landscape,
on which the non-elitist algorithm shows clear advantage over the elitist algorithms (see their hypervolume comparison in Table~\ref{tab:result}).
Figure~\ref{fig:nk20010} gives the solutions found by NE-MOEA and NSGA-II in a typical run on the NK-Landscape problem with $n=200$ and $k=10$.
As can be seen from the figure,
the solutions obtained by NE-MOEA not only have better diversity, 
but also significantly outperform the solutions obtained by NSGA-II in terms of convergence. 
The whole population of NSGA-II, which overlaps its archive, is dominated by many solutions in NE-MOEA's population, let alone the latter's archive.  
This indicates that NE-MOEA has better ability to explore the search space and jump out of local optima, even without a diversity maintenance mechanism.


Comparing the results amongst NSGA-II, NSGA-III and SMS-
EMOA in Table~\ref{tab:result}, we can see that the elitist algorithms
perform similarly. 
This implies that on combinatorial problems the selection criteria in the population update process may not matter very much as they all aim for preserving well-distributed nondominated solutions (thus the population may get stuck, even in very different areas, as recently reported in~\cite{Li2023}). 
In contrast, dominated solutions may play an important role in guiding the search to jump out of local optima.

\section{Conclusion}
This paper proposed a simple non-elitist MOEA (called NE-MOEA) that updates the evolutionary population with the newly-generated solutions only. 
NE-MOEA has demonstrated its competitiveness against mainstream elitist MOEAs on two combinatorial problems.
This is a very first study of non-elitist evolutionary multi-objective optimisation, 
and there are some limitations of the proposed NE-MOEA.
First, the algorithm requires a large population size and a large number of generations. 
Another limitation is that the algorithm is sensitive to variation rate (i.e., mutation rate here), although an upside is that the variation rate may possible be analytically obtained.

One subsequent piece of work is to study the potential of the non-elitist evolutionary search, for example, 
to investigate whether adding solutions' diversity information in their fitness and/or adding crossover variation can improve the proposed algorithm.
In addition,
testing them on more MOPs (e.g., well-established continuous MOPs), including real-world ones, can help further understand its performance and behaviour. 
Lastly,
a highly desirable future research direction is to attempt to conduct the theoretical analysis of the benefits of the non-elitist MOEAs, 
as did in non-elitist single-objective EAs~\cite{dang_non-elitist_2021, hevia_fajardo_self-adjusting_2021}.

\bibliographystyle{ACM-Reference-Format}


\end{document}